\documentclass[a4paper, 10pt, conference]{ieeeconf}      
\usepackage{FG2017}

\FGfinalcopy 

\IEEEoverridecommandlockouts                              
\usepackage{graphics} 
\usepackage{epsfig} 
\usepackage{amsmath} 
\usepackage{multirow}
\usepackage{hhline}

\title{\LARGE \bf
FERA 2017 - Addressing Head Pose in the Third Facial Expression Recognition and Analysis Challenge
}

\author{\parbox{16cm}{\centering
    {\large Michel F. Valstar$^1$, Enrique S\'anchez-Lozano$^1$, Jeffrey F. Cohn$^{2,6}$, L{\'a}szl{\'o} A. Jeni$^6$,\\Jeffrey M. Girard$^2$, Zheng Zhang$^3$, Lijun Yin$^3$, and Maja Pantic$^{4,5}$}\\
    {\normalsize
    $^1$ School of Computer Science, University of Nottingham, UK\\
    $^2$ Department of Psychology, University of Pittsburgh, Pittsburgh, USA\\
      $^3$ Department of Computer Science, Binghamton University, Binghamton, USA\\
       $^4$ Department of Computing, Imperial College London, London, UK\\
        $^5$  Electrical Engineering, Mathematics and Computer Science, University of Twente, The Netherlands\\
         $^6$ Robotics Institute, Carnegie Mellon University, Pittsburgh, USA}}
    \thanks{Support was provided in part by National Science Foundation awards CNS-1629898, CNS-1629716, CNS-1205664, CNS-1205195, IIS-1051103, and IIS-1051169, National Institutes of Health award MH 096951, and the European Union's Horizon 2020 research and innovation programme under grant agreement No. 645378.}
}

\begin{document}

\ifFGfinal
\thispagestyle{empty}
\pagestyle{empty}
\else
\author{Anonymous FG 2017 submission\\-- DO NOT DISTRIBUTE --\\}
\pagestyle{plain}
\fi
\maketitle

\begin{abstract}
The field of Automatic Facial Expression Analysis has grown rapidly in recent years. However, despite progress in new approaches as well as benchmarking efforts, most evaluations still focus on either posed expressions, near-frontal recordings, or both. This makes it hard to tell how existing expression recognition approaches perform under conditions where faces appear in a wide range of poses (or camera views), displaying ecologically valid expressions. The main obstacle for assessing this is the availability of suitable data, and the challenge proposed here addresses this limitation. The FG 2017 Facial Expression Recognition and Analysis challenge (FERA 2017) extends FERA 2015 to the estimation of Action Units occurrence and intensity under different camera views. In this paper we present the third challenge in automatic recognition of facial expressions, to be held in conjunction with the 12th IEEE conference on Face and Gesture Recognition, May 2017, in Washington, United States. Two sub-challenges are defined: the detection of AU occurrence, and the estimation of AU intensity. In this work we outline the evaluation protocol, the data used, and the results of a baseline method for both sub-challenges.

\end{abstract}

\section{INTRODUCTION}

\noindent Facial expression analysis is a rapidly growing field of research, due to the constantly increasing interest in, and feasibility of applying automatic human behaviour analysis to all kinds of multimedia recordings involving people. Applications include classical psychology studies, market research, interactions with virtual humans, multimedia retrieval, and the study of medical conditions that alter expressive behaviour \cite{Valstar2014_ABU}. Given the increasing prominence and utility of expression recognition systems, it is is increasingly important that such systems can be evaluated fairly and compared systematically. The FG 2017 Facial Expression Recognition and Analysis challenge (FERA 2017) shall support this effort by addressing three aspects frequently ignored in existing benchmarks: head-pose, expression intensity, and video duration.

Most Facial Expression Recognition and Analysis systems proposed in the literature focus on analysis of expressions from frontal faces. While it can be argued that in many scenarios people's faces will indeed be largely frontal most of the time, there are also many conditions in which either the camera angle is such that obtaining frontal views is unrealistic, or where the head pose with respect to the camera varies widely over time.

There are a few databases that include a number of non-frontal head-poses of a limited set of posed expressions. Multi-PIE recorded a very small number of expressions simultaneously with 15 cameras placed around the subject in a well-lit office setting \cite{GrossEtAl2010_MPI}. The MMI-Face database obtained frontal and profile views of all  Facial Action Coding System Action Units (FACS AUs, \cite{FACS2}) and six basic emotions using a mirror \cite{ValstarEtAl10_IDH}. But only databases recorded with depth-sensing cameras can generate an almost arbitrary set of face views of the same facial expression. One example database is the Bosphorus corpus \cite{SavranEtAl2008_BD3}. 

A second limitation of many existing benchmark databases is that they assume expression intensity is fixed, and as such they do not support the evaluation of intensity estimation. Most databases focus on detecting the occurrence of expressions, regardless of the significant differences in appearance, shape, and temporal dynamics caused by different expression intensities. In reality, expressions can vary greatly in intensity, and intensity is often a crucial cue for the interpretation of the meaning of expressions. 

Indeed McKeown et al. \cite{McKeown2015_UNL} have argued that the level of intensity is the key dimension in facial expressions that distinguishes whether they are delivered for socio-communicative functions at low levels of intensity or that they become hard-to-fake signals indicating that the expression is associated with a genuine felt emotion at high levels of intensity. If this is the case then intensity may be one of the most important features in assessing a user's psychological state from facial expressions.  However, very little annotated data is available for the evaluation of AU intensity estimation approaches. FERA 2015 made a significant step towards benchmarking AU intensity estimation, however, the data used in that challenge was predominantly of (near) frontal views \cite{ValstarEtAl2015_F2S}.

Finally, despite efforts towards evaluation standards of face video lasting longer than a few seconds (e.g. FERA 2011 \cite{ValstarEtAl2011_FFE}), video duration remains an issue that must be addressed by benchmarking challenges. In particular, the community needs to move away from evaluation procedures where each video recording consists of only a single expression, often with the onset and offset of an expression expressly defined. Instead, we need unsegmented videos that show multiple expression, with ideally expressions naturally transitioning one into another, without explicit neutral divisions in between.

In these respects, FERA 2017 shall help raise the bar for expression recognition by challenging participants to estimate AU intensity in face video of variable duration with unknown head-pose, thereby continuing to bridge the gap between excellent research on facial expression recognition and comparability and replication of results. In FERA 2017, we will use the BP4D+ dataset \cite{ZhangEtAl2016_MSE} to generate from every video 9 different 2D views, based the underlying 3D source data. The challenge is to detect the occurrence  and intensity of AUs, without knowing a priori what the facial view will be. We do this by means of two selected tasks: the detection of FACS Action Unit occurrence (Occurrence Detection Sub-Challenge), and fully automatic AU intensity estimation where the occurrence of AUS is not known beforehand (Intensity Estimation Sub-Challenge).
 
 \section{RELATED WORK \label{s:relatedwork}}

\begin{figure*}[t!]
  \centering \includegraphics[width=2.0\columnwidth]{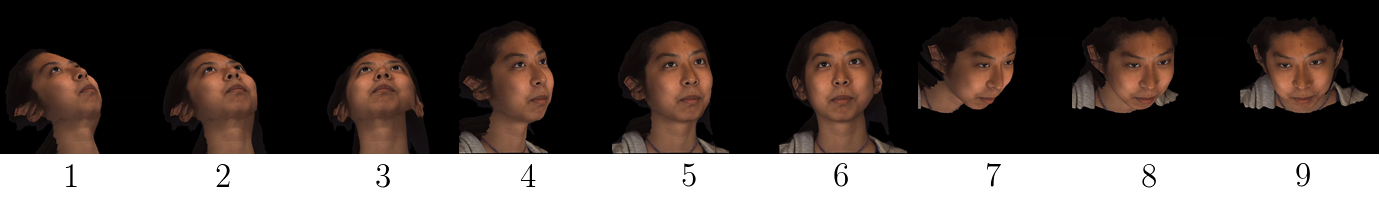}
  \caption{Each of the different views considered for the FERA 2017 Challenge.}
\label{f:views}
\end{figure*}

\noindent Facial expression recognition in general and action unit detection in particular have been studied extensively over the past decade. As a result, it is impossible to provide a comprehensive review of the field here. Instead we provide an overview of the relevant works only, focussing on methods that target AU occurrence detection and intensity estimation. For a general overview of the field of expression recognition we refer the reader to excellent recent surveys \cite{Corneanu2016, ZengEtAl09}.

\subsection{AU occurrence detection}
\noindent Common binary classifiers applied to this problem include Artificial Neural Networks (ANN), Boosting techniques, and Support Vector Machines (SVM). ANNs were the most popular method in earlier works (e.g. \cite{TianEtAl02}, \cite{BazzoLamar04}). Boosting algorithms, such as AdaBoost and GentleBoost, have been a common choice for AU recognition (e.g. \cite{HammEtAl11}, \cite{YangEtAl09}). Boosting algorithms are simple and quick to train. They have fewer parameters than SVM or ANN, and can be less prone to overfitting. They implicitly perform feature selection, which is desirable for handling high-dimensional data and speeding up inference, and can handle multiclass classification. SVMs are currently the most popular choice (e.g. \cite{ChewEtAl12}, \cite{WuEtAl12}, \cite{MahoorEtAl09}). SVMs provide good performance, can be non-linear, parameter optimisation is relatively easy, as efficient implementations are readily available, and a choice of kernel functions provides extreme flexibility of design.

\subsection{AU intensity estimation}
\noindent The goal in AU intensity estimation is to assign a per-frame label with possible integer value from 0 to 5 for each AU. This problem can be approached using either a classification or a regression learning method. 

\textit{Classification-based methods:} Some approaches use the confidence of a (binary) frame-based AU activation classifier to estimate  AU intensity. The rationale is that the lower the intensity is, the harder the classification will be. For example, Bartlett et al.  used the distance of the test sample to the SVM separating hyperplane \cite{BartlettEtAl06}, while Hamm et al.  used the confidence of the decision given by AdaBoost \cite{HammEtAl11}. 

It is however more natural to treat the problem as 6-class classification. For example, Mahoor et al.  employed six one-vs.-all binary SVM classifiers \cite{MahoorEtAl09}. Alternatively, a single multi-class classifier (e.g. ANN or a Boosting variant) could be used. The extremely large class overlap means however that such approaches are unlikely to be optimal. Girard \cite{Girard2014} found that multi-class and regression-based approaches more accurately detected intensity in comparison with distance-from-hyperplane based measures. 

\textit{Regression-based methods:} AU intensity estimation is nowadays often posed as a regression problem. Regression methods penalise incorrect labelling proportionally to the difference between ground truth and prediction. Such ordinal consideration of the labels is absent in classification methods. The large overlap between classes also implies an underlying continuous nature of intensity that regression techniques are better equipped to model. Examples include Support Vector Regression (\cite{JeniEtAl13}, \cite{Girard2014} and \cite{SavranEtAl11}). Kaltwang et al.   instead used Relevance Vector Regression to obtain a probabilistic prediction \cite{KaltwangEtAl12}. 

\section{DATA}
\noindent The training data for the FERA 2017 challenge is derived from the BP4D-Spontaneous database \cite{ZhangEtAl2014_BHS}, and the validation and test data of FERA 2017 is derived from a subset of BP4D+ database \cite{ZhangEtAl2016_MSE}. Data is split into train, validation, and test partitions. The train and development partitions are publicly available for researchers to train and develop their AU analysis systems, and to allow participants to uniformly report performance (i.e. using cross-validation). The test partition is held back by the organisers. Participants submit their trained systems and the FERA 2017 organisers apply their systems on this held-back data to create a fair comparison.

The challenge will focus on 10 AUs that occurred frequently in the BP4D dataset. The Occurrence Detection sub-challenge requires participants to detect 10 AUs from the BP4D database (see Table \ref{t:challengeAUs}).  AUs were selected based on their frequency of occurrence and sufficiently high inter-rater reliability scores. AU intensity estimation will be done on a subset of 7 AUs only (see Table \ref{t:challengeAUs}).


\subsection{Multiview Face Synthesis}
\label{ssec:mview}
\noindent Contrary to the FERA 2015 challenge, for FERA 2017, nine videos were created for each corresponding recording, ranging different face orientations, using the 3D models captured for each of the subjects. To create nine different face orientations, 3D sequences in BP4D and BP4D+ were rotated by -40, -20, and 0 degrees pitch and -40, 0, and 40 degrees yaw from frontal pose using ZFace \cite{Jeni16ImaVis_ZFace} and the known correspondence between 2D and 3D vertices. ZFace is real-time face alignment software that accomplishes dense 3D registration from 2D videos and images without requiring person-specific training.  We calculated the true 3D locations of the facial landmarks by mapping them to the ground truth 3D meshes. Faces were centred and scale was normalised to the average interocular distance of all subjects. An example of the resulting pose orientations addressed in this challenge is shown in Figure~\ref{f:views}.

\begin{table}
\begin{center}
\caption{Overview of AUs included in the two sub-challenges}
\label{t:challengeAUs}
\begin{tabular}{|l|c|c|}
\hline
 & \bf{Occurrence detection} & \bf{Intensity Estimation} \\
\hline
BP4D & AU1, AU4, AU6, AU7, AU10 & AU1, AU4, AU6, AU10  \\
& AU12, AU14, AU15, AU17, AU23 & AU12, AU14, AU17 \\
\hline
\end{tabular}
\end{center}
\end{table}

\subsection{BP4D Database}


\noindent Both the train and test partitions of the BP4D and BP4D+ databases consist of video data of young adults responding to emotion-elicitation tasks. The datasets are described in detail below. Here we note differences between them that are most relevant to the challenge. The training data was collected first and is publicly available \cite{ZhangEtAl2014_BHS}. The testing data is newer, part of the new collection \cite{ZhangEtAl2016_MSE} that includes 2D, 3D, thermal imaging and peripheral physiology, and will be released later. The number of participants in the two partitions is 41 and 20, respectively. Some differences exist in the threshold for coding AU occurrence and intensity, and changes occurred in the mix of AU coders of the two partitions. Coders were highly trained for both, and reliability was tested throughout coding to ensure consistency. 

The train partition of BP4D is selected from BP4D-Original \cite{ZhangEtAl2014_BHS}, and the test partition from BP4D-Expanded (a.k.a. BP4D+ \cite{ZhangEtAl2016_MSE}). Below we will refer to these as BP4D-Train and BP4D-Test.

\emph{BP4D-Train}
The BP4D-Train dataset includes digital video of 41 participants (56.1\% female, 49.1\% white, ages 18-29). These individuals were recruited from the departments of psychology and computer science and from the school of engineering at Binghamton University. All participants gave informed consent to the procedures and permissible uses of their data. Participants sat approximately 51 inches in front of a Di3D dynamic face capturing system during a series of eight emotion elicitation tasks.

To elicit target emotional expressions and conversational behaviour, we used approaches adapted from other investigators plus techniques that proved promising in pilot testing. Each task was administered by an experimenter who was a professional actor/director of performing arts. The procedures were designed to elicit a range of emotions and facial expressions that include happiness/amusement, sadness, surprise/startle, embarrassment, fear/nervous, physical pain, anger/upset, and disgust.

\emph{BP4D-Validation}
The BP4D-Validation dataset includes digital videos of 20 participants, which is a subset of BP4D+ \cite{ZhangEtAl2016_MSE}, with similar demographics as BP4D-original. It corresponds to the subjects belonging to the test set of FERA 2015. These individuals underwent similar recruitment, emotion-elicitation, and video recording procedures as those in the BP4D-Train dataset. The main difference between these datasets is that the extended dataset also collected physiological data and captured thermal images of participants. However, thermal and physiological data are not included in the FERA Challenge.

\emph{BP4D-Test}
The BP4D-Test includes digital videos of 30 participants, which was selected from a subset of BP4D+ \cite{ZhangEtAl2016_MSE}. 

In summary, there are 328 sessions from 41 subjects in the training, 159 sessions from 20 subjects in the development (a.k.a. validation), and 120 sessions from 30 subjects in the test partition. 
In total with 9 different views of each subject, there are 2952 videos in the training partition, 1431 videos in the development, and 1080 videos in test partition.



\subsection{Action Unit Annotation}
\noindent Action Units were annotated by a team of experts. Both databases were annotated frame-by-frame for the occurrence (i.e. activation) and intensity of AUs, using the Facial Action Coding System (FACS, \cite{FACS2}). FACS is a system for human observer coding of facial expressions, decomposing expressions into anatomically-based action units that correspond to specific facial muscles or muscle groups. Action units (AU) individually or in combinations can describe nearly all possible facial expressions. 

\emph{Occurrence Annotation}
\noindent For BP4D-Train, coders annotated onsets when AUs reached the A-level of intensity and offsets when they dropped below it. Segments of the most facially-expressive 20 seconds of each task were selected for coding. Across all participants, AU base occurrence rates, defined as the fraction of coded frames in which an AU occurred, averaged 35.4\%, and ranged from 17\% for to 59\%.
To assess inter-coder reliability, approximately 11\% of the data was independently coded by two highly trained and certified coders. Inter-coder reliability, as quantified by the Matthews Correlation Coefficient (MCC; \cite{Powers2011_EPR}), averaged 0.91. MCC for individual AU ranged from 0.81 for AU 23 to 0.96 for AU 2. These results suggest very strong inter-coder reliability for occurrence.  


For BP4D-Validation and BP4D-Test, coders annotated onsets when AUs reached the B-level of intensity and offsets when they dropped below it. Segments of the most facially-expressive 20 seconds of each task were selected for coding. Across all participants, AU base rates averaged 26.2\%, ranging from 5\% to 60\%.
To assess inter-coder reliability for occurrence, approximately 15\% of the data were independently comparison coded as above. Inter-coder reliability, as quantified by MCC, averaged 0.79, ranging from 0.69 to 0.91. These results indicate strong to very strong inter-rater reliability.
Across all AUs except for AU 15, inter-coder reliability for occurrence was lower in the expanded dataset than in the original dataset. These differences may be due in part to differences in threshold for determining occurrence (B-level versus A-level) and the addition of two coders in BP4D-Expanded (a.k.a BP4D+).

\emph{Intensity Annotation}
For BP4D-Original, seven AUs were intensity coded in the BP4D-Original dataset: AU1, AU4, AU6, AU10, AU12, AU14, and AU17. The B- and C-levels of intensity were most common for all except AU1, AU4, and AU 17, which showed more A- than C-level intensity.
To assess inter-coder reliability for intensity, approximately 6\% of the data was independently coded by two highly trained and certified coders. Inter-coder reliability, as quantified by the intra-class correlation coefficient (ICC; \cite{ShroutFleiss1979_ICU}), averaged 0.85. ICC for individual AU ranged from 0.79 to 0.92. These results indicate strong to very strong inter-coder reliability for intensity.



%
%

\section{EVALUATION PROCEDURE}
\noindent To perform a fair evaluation of participants' performance, participants are asked to submit their working programs to the challenge organisers, who will run these programs on the held-back test set. 

The evaluation will be view-independent. Participants' working programs will be evaluated indistinctly in all the videos in the test set, and no prior information of the specific view will be given. As such, participants should consider each of the views as independent videos, although for each user and task, the annotations corresponding to the 9 views will all be the same. 

The performance measure for AU occurrence is the F1-measure, which is the harmonic mean of recall and precision. For an AU with precision $P$ and recall $R$, it is calculated as:

\begin{equation}
F_1=\frac{2PR}{P+R}
\end{equation}

The performance measure for AU intensity is the Intraclass Correlation Coefficient (ICC, \cite{ShroutFleiss1979_ICU}). Given ground truth labels $\mathbf{y}$, $y_t \in \{0, 1, ... 5\}$ and predictions $\mathbf{\hat{y}}$, $\hat{y}_t \in \mathcal{N}$, the ICC $I$ is calculated as follows: 


\begin{equation}
I = \frac{W-S}{W+(k-1)W}
\end{equation}

\noindent where $k$ is the number of coding sources compared; in our case $k=2$.  $W$ and $S$ are the Within-target Mean Squares and Residual Sum of Squares, respectively, and are computed as follows:

\begin{equation}
W = \sum_{i=1}^n\sum_{j=1}^k\frac{(y_{ij}-\bar{y_i})^2}{n(k-1)} =  \sum_{i=1}^n\sum_{j=1}^k\frac{(y_{ij}-\bar{y_i})^2}{n} \label{e:WMS}
\end{equation}

\noindent where $\bar{y_i} = \sum_{j=1}^ky_{ij}/k$ and the third term of Eq. (\ref{e:WMS}) follows from $k=2$. $S$ is defined as:


\begin{equation}
S= \sum_{i=1}^n(y_i-\hat{y}_i)^2
\end{equation}


To come to a single score $s$ for the Occurrence Detection and Intensity Estimation Sub-Challenges, labels of all test sequences will be concatenated into a sequence to calculate F1/ICC measures per AU. The average value will be used as the performance of a participant's submission:

\begin{equation}
s = \frac{1}{N}\sum_{a=1}^Nf_a(\mathbf{y},\mathbf{\hat{y}})
\end{equation}

\noindent where $f_a$ is either the F1 or ICC measure for a given AU $a$, depending on the sub-challenge, and $N$ is either the 10 AUs for Occurrence Detection, or the 7 AUs for intensity estimation. 


For the baseline system, a number of different performance measures are shown, given that each has its own merits, and combined they provide a deeper analysis of the results. For the Occurrence detection, the measures are the F1, the Accuracy, and the 2AFC score. Whereas F1 and Accuracy are well-known performance measurement, 2AFC is less well-known. The 2AFC score is a good approximation of the area under the receiver operator characteristic curve (AUC). In contrast to F1, 2AFC does take True Negative preditions into account. In this study the 2AFC has been calculated based on the CRF/CORF (see Section~\ref{ssec:baseline_results} for more details) likelihood values as follows:
	\begin{align}
		2AFC(\hat{Y}) = \sum^n_{i = 0}\sum^p_{j = 0}\sigma(P_j, N_i)\frac{1}{n \times p},
	\end{align}
	$$		
		\sigma(X, Y) =
			\begin{cases}
				1, & \mbox{if } X > Y \\
				0.5, & \mbox{if } X == Y \\
				0, & \mbox{if } X < Y
			\end{cases} 
	$$
where $\hat{Y}$ is a vector of decision function output values, $n$ is the total number of true negative and $p$ the total number of true positive instances in $\hat{Y}$, and $P$ and $N$ are subsets of $\hat{Y}$ corresponding to all positive and negative instances, respectively. 

For the Intensity sub-challenge, the used measures are the Root Mean Squared Error (RMSE), the Intra-Class Correlation (ICC), and the Pearson Correlation Coefficient (PCC).

\section{BASELINE SYSTEM}
\noindent In this work we provide baseline recognition results on both the development and test sets, for easy comparison of participants' systems. Contrary to previous challenges, in FERA 2017 the baseline points and features are not made publicly available, as participants are asked to submit their methods to run as standalone applications, in which all challenging tasks associated to the different views must be addressed by participants.

\subsection{Baseline Features}
\begin{figure}[h!]
	\vspace{-10pt}
	\begin{center}
		\includegraphics[width=0.80\columnwidth]{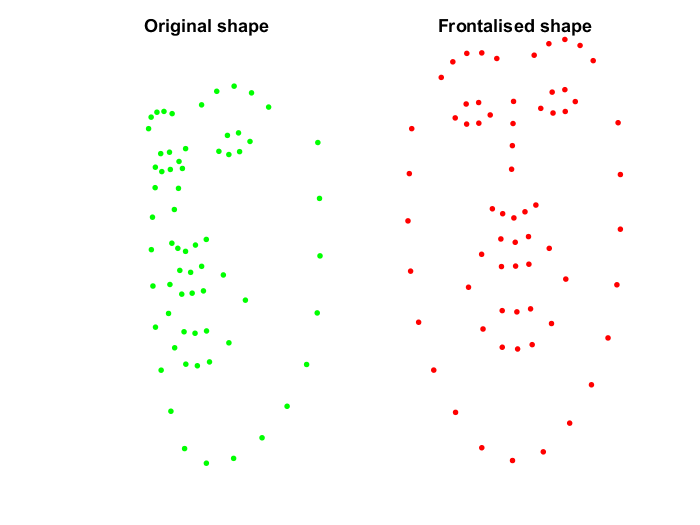}
	\end{center}
	\vspace{-20pt}
	\caption{Left image corresponds to a shape captured during tracking, and the right image corresponds to the normalised and frontalised shape. It can be seen that the main expression remains whilst pose has been removed}
	\label{fig:shapes}
\end{figure}

\noindent For the challenge baseline we used geometric features derived from tracked facial point locations. The geometric features are based on the 66 landmarks detected and subsequently tracked with the Cascaded Continuous Regression facial point detector/tracker proposed by S\'anchez-Lozano et al. \cite{Sanchez16ECCV,Sanchez16arXiv}. In some cases the facial point detection failed, and therefore the points are simply set to 0. In order to compute the features the points are registered with respect to the tracker's Shape Model. The Shape Model consists of 24 parameters, the first four of which correspond to rigid information, whereas the last 20 correspond to pose and expression, and was built using the training partition of the 300VW database~\cite{shen15}, using both the original annotations, and the mirrored points. This way, the first non-rigid shape parameter is responsible of encoding only the pose angle~\cite{gonzalez07}. In order to unify the feature extraction and make it view-independent, the pose-related parameter is set to zero, and then the 19 non-rigid parameters corresponding to expression are taken as the low-dimensional representation of the shape. 
Figure~\ref{fig:shapes} shows an example of a posed-face and the reconstructed shape using the 19 non-rigid parameters. Reconstructing the face using only the last 19 non-rigid parameters, and encoding the rigid parameters to zero, implies that faces are automatically registered and normalised, and pose is removed. These 19 features represent the PCA parameters that would result after applying a dimensionality reduction to the registered and normalised points. 
Then, a set of geometric features is extracted from the reconstructed shapes. The first 19 features are the non-rigid parameters described above. The next 19 features are composed by subtracting the parameters of the previous frame from that of the current one\footnote{Note that subtracting the shape parameters is directly equivalent to subtracting the reconstructed shapes themselves}. This applies to all frames except the very first one of every session, for which these features are the same as the first 19. 
For the next set of features the 49 inner facial landmarks have been split into three groups representing the left eye (points 20 - 25) and left eyebrow (points 1 - 5), the right eye (points 26 - 31) and right eyebrow (points 6 - 10), and the mouth region (points 32 - 49). For each of these groups a set of features representing Euclidean distances as well as angles in radians between points within the groups is extracted.

Distances between points within a group are computed by taking the squared L2-norm between consecutive points:
$$F(i) = \|\tilde{p}_i - \tilde{p}_{i+1}\|_2^2,$$
$$i = \{1..N_p-1\}$$
where $N_p$ is the total number of points within the region, $\tilde{p}_i$ is the point coordinates vector and $F$ is the feature array of the region. Hence, for each group the number of features constructed in this manner $N_f$ is equal to:
$$N_f = N_p - 1$$

The same approach is used to calculate the angles between two lines defined by two pairs of points at a time within a group, where the two pairs share one common point. For each consecutive triplet of points Euclidean distances between them are computed first, which are then used to calculate angle between the points: 
$$F(i) = \arccos\left(\frac{\tilde{p}_{12}^2 + \tilde{p}_{13}^2 - \tilde{p}_{23}^2}{2 * \tilde{p}_{12} * \tilde{p}_{13}}\right)$$
where $\tilde{p}_{ij}$ is an Euclidean distance between points $i$ and $j$. The number of features extracted this way is equal to the total number of consecutive angles within the groups of points, which is equal to:
$$N_f = N_p - 2$$
There are thus 71 features in total extracted from the above face regions. 

Finally, for the last 49 features we first compute median of a set of stable points of the aligned shape, meant to be robust to change in pose and expression. We then go through all of the aligned shape points and compute Euclidean distance between them and the median. In total there are 158 geometric features extracted from every video frame in the database.

\begin{table*}[t]
\begin{center}
\caption{Baseline results for the occurrence sub-challenge on the development and test partition measured in F1 score, 2AFC and Accuracy.}
\vspace{2mm}
\label{t:baseline_occurrence_results}
\begin{tabular}{c|c|c|c|c|c|c}
\hline
\multirow{2}{*}{Action Unit} & \multicolumn{3}{c|}{Development} & \multicolumn{3}{c}{Test} \\
\hhline{~------}
& F1 & 2AFC & Accuracy & F1 & 2AFC & Accuracy \\
\hline
AU1 & 0.154 & 0.560 & 0.570 & 0.147 & 0.543 & 0.530 \\  
AU4 & 0.172 & 0.510 & 0.520 & 0.044 & 0.488 & 0.557 \\ 
AU6 & 0.564 & 0.473 & 0.676 & 0.630 & 0.488 & 0.662 \\
AU7 & 0.727 & 0.550 & 0.642 & 0.755 & 0.579 & 0.664 \\  
AU10 & 0.692 & 0.649 & 0.638 & 0.758 & 0.684 & 0.671 \\ 
AU12 & 0.647 & 0.547 & 0.660 & 0.687 & 0.566 & 0.651 \\  
AU14 & 0.622 & 0.507 & 0.622 & 0.668 & 0.523 & 0.615 \\ 
AU15 & 0.146 & 0.492 & 0.307 & 0.220 & 0.494 & 0.310 \\ 
AU17 & 0.224 & 0.506 & 0.485 & 0.274 & 0.503 & 0.522 \\
AU23 & 0.207 & 0.496 & 0.373 & 0.342 & 0.498 & 0.432 \\ 
\hline
Mean & 0.416 & 0.529 & 0.549 & 0.452 & 0.537 & 0.561 \\
\hline
\end{tabular}
\end{center}
\end{table*}

\begin{table*}[t]
\begin{center}
\caption{Baseline results for the occurrence sub-challenge on the development partition, per view}
\vspace{2mm}
\label{t:baseline_occurrence_per_view}
\begin{tabular}{c|c|c|c|c|c|c|c|c|c|}
\hhline{----------} 
& \multicolumn{9}{c|}{F1 score} \\
\hline
View & 1 & 2 & 3 & 4 & 5 & 6 & 7 & 8 & 9 \\
\hline
\% Detected frames & 66.87 &  98.18 & 99.99 & 99.96 & 100 & 100 & 88.67 & 98.14 & 99.46 \\
\hline
AU1 & 0.103 & 0.150 & 0.136 & 0.193 & 0.196 & 0.171 & 0.180 & 0.145 & 0.123 \\ 
AU4 & 0.150 & 0.159 & 0.148 & 0.191 & 0.190 & 0.183 & 0.202 & 0.202 & 0.175 \\ 
AU6 & 0.505 & 0.557 & 0.134 & 0.689 & 0.747 & 0.724 & 0.560 & 0.532 & 0.493 \\ 
AU7 & 0.721 & 0.729 & 0.413 & 0.746 & 0.797 & 0.787 & 0.716 & 0.747 & 0.758 \\ 
AU10 & 0.554 & 0.710 & 0.642 & 0.777 & 0.776 & 0.750 & 0.639 & 0.679 & 0.659 \\ 
AU12 & 0.522 & 0.678 & 0.184 & 0.786 & 0.809 & 0.771 & 0.596 & 0.638 & 0.601 \\ 
AU14 & 0.515 & 0.563 & 0.090 & 0.675 & 0.724 & 0.744 & 0.619 & 0.670 & 0.678 \\ 
AU15 & 0.131 & 0.150 & 0.142 & 0.146 & 0.146 & 0.146 & 0.143 & 0.159 & 0.152 \\ 
AU17 & 0.173 & 0.251 & 0.235 & 0.242 & 0.246 & 0.241 & 0.220 & 0.211 & 0.195 \\ 
AU23 & 0.227 & 0.229 & 0.199 & 0.208 & 0.196 & 0.166 & 0.201 & 0.201 & 0.208 \\ 
\hline
Mean & 0.360 & 0.418 & 0.232 & 0.465 & 0.482 & 0.468 & 0.408 & 0.418 & 0.404 
 \\ \hline
& \multicolumn{9}{c|}{Accuracy} \\
\hline
View & 1 & 2 & 3 & 4 & 5 & 6 & 7 & 8 & 9 \\
\hline
AU1 & 0.252 & 0.359 & 0.329 & 0.563 & 0.580 & 0.500 & 0.729 & 0.902 & 0.915 \\ 
AU4 & 0.300 & 0.372 & 0.143 & 0.456 & 0.554 & 0.627 & 0.501 & 0.810 & 0.920 \\ 
AU6 & 0.732 & 0.727 & 0.696 & 0.813 & 0.829 & 0.791 & 0.667 & 0.473 & 0.357 \\ 
AU7 & 0.606 & 0.642 & 0.539 & 0.695 & 0.738 & 0.728 & 0.610 & 0.604 & 0.617 \\ 
AU10 & 0.563 & 0.650 & 0.711 & 0.749 & 0.722 & 0.675 & 0.612 & 0.560 & 0.503 \\ 
AU12 & 0.627 & 0.707 & 0.609 & 0.810 & 0.811 & 0.764 & 0.636 & 0.539 & 0.436 \\ 
AU14 & 0.618 & 0.626 & 0.516 & 0.712 & 0.730 & 0.730 & 0.608 & 0.533 & 0.524 \\ 
AU15 & 0.398 & 0.218 & 0.102 & 0.678 & 0.215 & 0.121 & 0.368 & 0.285 & 0.378 \\ 
AU17 & 0.699 & 0.468 & 0.246 & 0.741 & 0.598 & 0.522 & 0.464 & 0.306 & 0.323 \\
AU23 & 0.271 & 0.275 & 0.554 & 0.508 & 0.466 & 0.594 & 0.349 & 0.189 & 0.154 \\ 
\hline
Mean & 0.507 & 0.505 & 0.444 & 0.673 & 0.624 & 0.605 & 0.554 & 0.520 & 0.513 
 \\
\hline
\end{tabular}
\end{center}
\end{table*}

\begin{table*}[t]
\begin{center}
\caption{Baseline results for the occurrence sub-challenge on the test partition, per view}
\vspace{2mm}
\label{t:baseline_occurrence_per_view_test}
\begin{tabular}{c|c|c|c|c|c|c|c|c|c|}
\hhline{----------} 
& \multicolumn{9}{c|}{F1 score} \\
\hline
View & 1 & 2 & 3 & 4 & 5 & 6 & 7 & 8 & 9 \\
\hline
\% Detected frames & 73.27 &  98.97 & 100 & 99.92 & 100 & 100 & 85.31 & 99.19 & 99.86 \\
\hline
AU1 & 0.114 & 0.147 & 0.154 & 0.160 & 0.189 & 0.173 & 0.118 & 0.035 & 0.005 \\ 
AU4 & 0.044 & 0.043 & 0.055 & 0.041 & 0.032 & 0.037 & 0.047 & 0.041 & 0.015 \\ 
AU6 & 0.550 & 0.602 & 0.204 & 0.704 & 0.808 & 0.787 & 0.565 & 0.643 & 0.615 \\ 
AU7 & 0.742 & 0.770 & 0.329 & 0.788 & 0.837 & 0.840 & 0.758 & 0.777 & 0.779 \\ 
AU10 & 0.660 & 0.772 & 0.670 & 0.839 & 0.831 & 0.812 & 0.663 & 0.767 & 0.758 \\ 
AU12 & 0.609 & 0.702 & 0.261 & 0.811 & 0.800 & 0.798 & 0.609 & 0.698 & 0.678 \\ 
AU14 & 0.593 & 0.611 & 0.109 & 0.663 & 0.776 & 0.803 & 0.631 & 0.745 & 0.749 \\ 
AU15 & 0.162 & 0.232 & 0.227 & 0.147 & 0.225 & 0.229 & 0.230 & 0.236 & 0.237 \\ 
AU17 & 0.164 & 0.346 & 0.306 & 0.177 & 0.306 & 0.311 & 0.229 & 0.260 & 0.266 \\ 
AU23 & 0.373 & 0.371 & 0.313 & 0.278 & 0.317 & 0.268 & 0.316 & 0.366 & 0.368 \\ 
\hline
Mean & 0.401 & 0.460 & 0.263 & 0.461 & 0.512 & 0.506 & 0.416 & 0.457 & 0.447 
 \\ \hline
& \multicolumn{9}{c|}{Accuracy} \\
\hline
View & 1 & 2 & 3 & 4 & 5 & 6 & 7 & 8 & 9 \\
\hline
AU1 & 0.251 & 0.331 & 0.403 & 0.361 & 0.505 & 0.438 & 0.641 & 0.912 & 0.932 \\ 
AU4 & 0.399 & 0.314 & 0.185 & 0.499 & 0.662 & 0.698 & 0.460 & 0.857 & 0.936 \\ 
AU6 & 0.637 & 0.700 & 0.601 & 0.779 & 0.832 & 0.805 & 0.621 & 0.526 & 0.456 \\ 
AU7 & 0.614 & 0.662 & 0.459 & 0.746 & 0.786 & 0.790 & 0.643 & 0.640 & 0.639 \\ 
AU10 & 0.595 & 0.678 & 0.656 & 0.786 & 0.754 & 0.718 & 0.587 & 0.643 & 0.619 \\ 
AU12 & 0.605 & 0.700 & 0.548 & 0.795 & 0.760 & 0.756 & 0.598 & 0.577 & 0.523 \\ 
AU14 & 0.583 & 0.607 & 0.423 & 0.652 & 0.736 & 0.757 & 0.581 & 0.599 & 0.599 \\ 
AU15 & 0.450 & 0.213 & 0.139 & 0.620 & 0.255 & 0.170 & 0.359 & 0.245 & 0.336 \\ 
AU17 & 0.715 & 0.564 & 0.353 & 0.753 & 0.665 & 0.559 & 0.476 & 0.249 & 0.361 \\ 
AU23 & 0.325 & 0.275 & 0.533 & 0.581 & 0.574 & 0.677 & 0.396 & 0.283 & 0.242 \\ 
\hline
Mean & 0.517 & 0.504 & 0.430 & 0.657 & 0.653 & 0.637 & 0.536 & 0.553 & 0.564 
 \\
\hline
\end{tabular}
\end{center}
\end{table*}


\begin{table*}[t]
\begin{center}
\caption{Baseline results for the intensity sub-challenge on the development and test partition measured in RMSE, PCC and ICC.}
\vspace{2mm}
\label{t:baseline_intensity_results}
\begin{tabular}{c|c|c|c|c|c|c}
\hline
\multirow{2}{*}{Action Unit} & \multicolumn{3}{c|}{Development} & \multicolumn{3}{c}{Test} \\
\hhline{~------}
& RMSE & PCC & ICC & RMSE & PCC & ICC \\
\hline
AU1 & 1.006 & 0.097 & 0.082 & 1.082 & 0.040 & 0.035 \\ 
AU4 & 1.296 & 0.084 & 0.069 & 1.200 & -0.007 & -0.004 \\
AU6 & 1.648 & 0.429 & 0.429 & 1.604 & 0.463 & 0.461 \\ 
AU10 & 1.628 & 0.435 & 0.434 & 1.548 & 0.462 & 0.451 \\
AU12 & 1.345 & 0.543 & 0.540 & 1.339 & 0.518 & 0.518 \\ 
AU14 & 1.637 & 0.264 & 0.259 & 1.422 & 0.046 & 0.037 \\  
AU17 & 1.256 & 0.052 & 0.005 & 1.626 & 0.024 & 0.020 \\ 
\hline
Mean & 1.402 & 0.265 & 0.260 & 1.403 & 0.221 & 0.217 \\
\hline
\end{tabular}
\end{center}
\end{table*}


\begin{table*}[t]
\begin{center}
\caption{Baseline results for the intensity sub-challenge on the development partition, per view}
\vspace{2mm}
\label{t:baseline_intensity_per_view}
\begin{tabular}{c|c|c|c|c|c|c|c|c|c|c|}
\hhline{-----------} 
 \multicolumn{2}{c|}{\% Detected frames} & 66.87 &  98.18 & 99.99 & 99.96 & 100 & 100 & 88.67 & 98.14 & 99.46 \\
\hline
& \multicolumn{10}{c|}{RMSE} \\
\hline
View & Chance & 1 & 2 & 3 & 4 & 5 & 6 & 7 & 8 & 9 \\
\hline
AU1 & 0.485 & 1.005 & 1.019 & 0.958 & 0.871 & 0.772 & 0.962 & 1.069 & 0.924 & 1.369 \\
AU4 & 0.532 & 1.246 & 1.416 & 1.287 & 1.090 & 1.012 & 1.028 & 1.431 & 1.630 & 1.390 \\
AU6 & 1.633 & 1.526 & 1.728 & 1.402 & 1.414 & 1.469 & 1.731 & 1.693 & 1.780 & 1.998 \\
AU10 & 1.913 & 1.646 & 1.720 & 1.477 & 1.246 & 1.346 & 1.607 & 1.624 & 1.616 & 2.192 \\
AU12 & 1.860 & 1.612 & 1.353 & 1.329 & 1.048 & 1.100 & 1.302 & 1.466 & 1.245 & 1.542 \\
AU14 & 1.923 & 1.735 & 1.620 & 1.693 & 1.616 & 1.550 & 1.496 & 1.679 & 1.584 & 1.742 \\
AU17 & 0.793 & 0.948 & 1.255 & 1.591 & 0.847 & 0.972 & 0.929 & 1.122 & 1.399 & 1.859 \\
\hline
Mean & 1.306 &1.388 & 1.444 & 1.391 & 1.162 & 1.174 & 1.294 & 1.441 & 1.454 & 1.727 \\ \hline
& \multicolumn{10}{c|}{ICC} \\
\hline
View & Chance & 1 & 2 & 3 & 4 & 5 & 6 & 7 & 8 & 9 \\
\hline
AU1& - & 0.017 & -0.037 & -0.001 & 0.196 & 0.263 & 0.200 & 0.073 & 0.085 & 0.018 \\ 
AU4 & - & 0.061 & 0.080 & 0.075 & 0.082 & 0.125 & 0.072 & 0.111 & 0.027 & 0.028 \\ 
AU6 & - & 0.322 & 0.324 & 0.396 & 0.642 & 0.630 & 0.584 & 0.375 & 0.454 & 0.280 \\ 
AU10 & - & 0.332 & 0.433 & 0.398 & 0.613 & 0.633 & 0.598 & 0.380 & 0.471 & 0.277 \\ 
AU12 & - & 0.286 & 0.463 & 0.506 & 0.768 & 0.778 & 0.756 & 0.432 & 0.583 & 0.317 \\ 
AU14 & - & 0.211 & 0.246 & 0.185 & 0.318 & 0.342 & 0.354 & 0.232 & 0.305 & 0.216 \\ 
AU17 & - & -0.018 & 0.048 & 0.065 & 0.023 & -0.022 & -0.011 & -0.023 & -0.009 & -0.032 \\ 
\hline
Mean & - & 0.173 & 0.222 & 0.232 & 0.377 & 0.392 & 0.365 & 0.226 &0.274 & 0.158
 \\
\hline
\end{tabular}
\end{center}
\end{table*}

\begin{table*}[t]
\begin{center}
\caption{Baseline results for the intensity sub-challenge on the test partition, per view}
\vspace{2mm}
\label{t:baseline_intensity_per_view_test}
\begin{tabular}{c|c|c|c|c|c|c|c|c|c|c|}
\hhline{-----------}  
\multicolumn{2}{c|}{\% Detected frames} & 66.87 &  98.18 & 99.99 & 99.96 & 100 & 100 & 88.67 & 98.14 & 99.46 \\ \hline
& \multicolumn{10}{c|}{RMSE} \\
\hline
View & Chance & 1 & 2 & 3 & 4 & 5 & 6 & 7 & 8 & 9 \\
\hline
AU1 & 0.523 & 1.036 & 0.875 & 0.955 & 1.233 & 1.044 & 1.245 & 1.046 & 0.946 & 1.278 \\
AU4 & 0.308 & 1.073 & 1.364 & 1.124 & 0.886 & 0.847 & 0.845 & 1.377 & 1.493 & 1.528 \\
AU6 & 1.641 & 1.642 & 1.553 & 1.315 & 1.518 & 1.502 & 1.609 & 1.715 & 1.706 & 1.824 \\
AU10 & 1.759 & 1.668 & 1.642 & 1.343 & 1.126 & 1.330 & 1.483 & 1.660 & 1.504 & 2.009 \\
AU12 & 1.765 & 1.606 & 1.217 & 1.334 & 1.203 & 1.255 & 1.344 & 1.519 & 1.215 & 1.296 \\
AU14 & 0.634 & 1.170 & 1.669 & 1.098 & 1.117 & 1.319 & 1.440 & 1.264 & 1.659 & 1.851 \\
AU17 & 0.665 & 1.199 & 1.358 & 2.216 & 1.107 & 1.186 & 1.324 & 1.756 & 2.135 & 1.889 \\
\hline
Mean & 1.042 & 1.342 & 1.383 & 1.341 &  1.170 & 1.212 & 1.327 & 1.477 & 1.523 & 1.668
 \\ \hline
& \multicolumn{10}{c|}{ICC} \\
\hline
View & Chance & 1 & 2 & 3 & 4 & 5 & 6 & 7 & 8 & 9 \\
\hline
AU1 & - & 0.006 & 0.104 & 0.087 & -0.014 & 0.055 & 0.025 & -0.020 & 0.035 & 0.060 \\ 
AU4 & - & -0.017 & -0.002 & 0.004 & 0.003 & -0.012 & 0.003 & -0.028 & -0.010 & 0.018 \\ 
AU6 & - & 0.293 & 0.464 & 0.437 & 0.613 & 0.628 & 0.616 & 0.309 & 0.529 & 0.409 \\ 
AU10 & - & 0.232 & 0.488 & 0.419 & 0.646 & 0.663 & 0.662 & 0.233 & 0.574 & 0.378 \\ 
AU12 & - & 0.244 & 0.540 & 0.496 & 0.675 & 0.706 & 0.709 & 0.275 & 0.601 & 0.512 \\ 
AU14 & - & 0.046 & 0.039 & 0.046 & -0.019 & 0.050 & 0.066 & -0.031 & 0.063 & 0.066 \\ 
AU17 & - & -0.027 & 0.035 & 0.092 & 0.055 & 0.019 & 0.015 & -0.023 & 0.034 & 0.014 \\ 
\hline
Mean & - & 0.111 & 0.238 & 0.226 & 0.280 & 0.301 & 0.299 & 0.102 &0.261 & 0.208  \\
\hline
\end{tabular}
\end{center}
\end{table*}

\subsection{Baseline Results}
\label{ssec:baseline_results}
\noindent The baseline system is kept simple on purpose since it should be easy to interpret and simple to replicate. Contrary to previous challenge editions, we decided to model the temporal dynamics, with the aim of covering misaligned frames. The learning method for the temporal dynamics modelling is Conditional Random Field (CRF, \cite{lafferty01}), for the AU occurrence sub-challenge and Conditional Ordinal Random Field (CORF, \cite{kim10}), for the intensity sub-challenge using the geometric features extracted using the method described above. 

We have divided the training videos into segments of 90 frames with a stride window of 30 frames. This way, given that expressions are spontaneous, we can encode short sequences. Moreover, short segments of missing frames can be compensated by the dynamics predicted by a graphical model, something that can not be done using a simple frame-by-frame estimator, which would be highly affected by the tracking system. Also, to avoid the influence of inaccurate tracked points for model training, and in order to prove the generalisation of our method to different views, we used for training only the geometric features extracted in the videos corresponding to views 5 and 6 (i.e., containing frontal faces). For each of the AUs, the training set is balanced so that the amount of samples per class is approximately the same (when possible). The dimensionality of the feature vector is 158, although it is reduced using a Correlation Feature Selection (CFS) method.

At test time, videos are evaluated following the same process: 90-frames windows are evaluated by the CRF/CORF model, with a stride of 30 frames. For each of the windows, inference returns the likelihood of the chosen class. Then, we have three predictions per-frame, with their corresponding likelihoods. The final per-frame assignment is given by the prediction attached to the maximum likelihood. 

Baseline results for occurrence/activation detection, and intensity estimation, are shown in Table~\ref{t:baseline_occurrence_results} and Table~\ref{t:baseline_intensity_results} respectively, for the development and test partitions. Detection performance is measured by F1 as well as Accuracy and 2AFC scores, for the occurrence challenge, and by ICC, RMSE and PCC scores for the intensity estimation challenge. A number of different performance measures are shown since each has their own merits, and combined they provide a deeper analysis of the results. However the challenge participants will only be judged based on F1/ICC scores. 

In order to demonstrate and evaluate the generalisation capabilities of the baseline system to different views, we have also included F1 scores and Accuracy per view, which are shown in Table~\ref{t:baseline_occurrence_per_view} and Table~\ref{t:baseline_occurrence_per_view_test}. The first row shows the percentage of frames per view that were properly detected/tracked. However, a high percentage does not necessarily mean that the accuracy of detected points is good enough to encode facial expressions. From the results shown in Table~\ref{t:baseline_occurrence_per_view}, and after visual inspection, it can be seen that videos corresponding to view 4 have results close to those of views 5 and 6, despite view 4 being non-frontal. Given that the face tracker performs well in these sequences, the frontalisation approach serves to achieve a good performance, especially given that none of the videos corresponding to view 4 were used to train the models. However, other extreme views were harder to track, and inaccurate point localisations were given, thus affecting the system's performance.  
In addition, it can be seen that the frontalisation also yields good results for the Intensity subchallenge. Results per view (RMSE and ICC) are shown in Table~\ref{t:baseline_intensity_per_view} and Table~\ref{t:baseline_intensity_per_view_test}, in which also the chance level is shown, understood as the error that is given by a naive classifier returning always the class that has the highest frequency in the training set (0 in all of them). Results show to outperform chance level, giving reasonable results. However, for AU1, AU4, and AU17, the amount of frames labelled with intensity zero is around the $90\%$ for the development set. This makes it hard for a graphical model to be accurate \footnote{In general, if a CRF/CORF is trained with a highly imbalanced number of training instances per class, then it is most likely to approach a naive classifier, whereas when balancing the training data, it is more likely to be less accurate}. This explains the low MSE measured, as well as the low ICC level. 

Despite the good results given for some of the views, there is still a huge gap to improve, especially for challenging views, such as view 1 and 9.

\section{CONCLUSION}
\noindent In this paper we have presented the Third Facial Expression Recognition and Analysis Challenge (FERA 2017) dedicated to FACS Action Units detection and intensity estimation on the highly challenging set of data. The dataset for this challenge has been derived from the BP4D, and extended to generate an extensive set of videos comprising 9 different views. This is the first time that a FACS AU annotated dataset is focused on expression analysis under different camera views, ranging extreme poses. The challenge addresses such significant problems of the field as expression intensity estimation as well as robust detection under non-frontal head poses, or partial self-occlusions. Baseline results obtained using geometric features demonstrate a huge room for potential improvements to be brought by the challenge participants, especially corresponding to challenging views.   

\bibliographystyle{abbrv}


\end{document}